\pdfoutput=1

\documentclass[11pt]{article}

\usepackage[]{EMNLP2022}

\usepackage{times}
\usepackage{latexsym}

\usepackage[T1]{fontenc}

\usepackage[utf8]{inputenc}

\usepackage{microtype}

\usepackage{inconsolata}

\usepackage{microtype}
\usepackage{graphicx}
\usepackage{multirow}
\usepackage{multicol}
\usepackage{microtype}
\usepackage{subcaption}
\usepackage{enumitem,kantlipsum}
\usepackage[normalem]{ulem}
\usepackage{tabularx}
\usepackage{tcolorbox}
\usepackage{color, colortbl}
\usepackage{todonotes} 

\usepackage{pbox}
\usepackage{booktabs} 
\usepackage{amssymb}
\usepackage{pifont}

\usepackage{tabu}
\usepackage{fontawesome}
\newcommand{\sq}{\faCheckSquare}

\newcommand{\rashkincorpus}{\texttt{TSHP-17}}
\newcommand{\proppycorpus}{\texttt{QProp} }

\newcommand{\rank}[2]{{\scriptsize \texttt{#1:#2}}}

\title{Overview of the WANLP 2022 Shared Task \\on Propaganda Detection in Arabic}

  \author{Firoj Alam$^1$, Hamdy Mubarak$^1$, Wajdi Zaghouani$^2$,  \\
  \textbf{Giovanni Da San Martino$^3$,
  Preslav Nakov$^4$}\\
  $^1$Qatar Computing Research Institute, HBKU, Qatar \\
  $^2$Hamad Bin Khalifa University, Qatar \\
  $^3$University of Padova, Italy\\
  $^4$Mohamed bin Zayed University of Artificial Intelligence, UAE\\  
  {\small \texttt{\{falam,hmubarak,wzaghouani\}@hbku.edu.qa},  \texttt{dasan@math.unipd.it}, \texttt{preslav.nakov@mbzuai.ac.ae}}
  \\}

\begin{document}
\maketitle
\begin{abstract}
Propaganda is the expression of an opinion or an action by an individual or a group deliberately designed to influence the opinions or the actions of other individuals or groups with reference to predetermined ends, which is achieved by means of well-defined rhetorical and psychological devices. Propaganda techniques are commonly used in social media to manipulate or to mislead users. Thus, there has been a lot of recent research on automatic detection of propaganda techniques in text as well as in memes. However, so far the focus has been primarily on English. With the aim to bridge this language gap, we ran a \textit{shared task on detecting propaganda techniques in Arabic tweets} as part of the WANLP 2022 workshop, which included two subtasks. Subtask~1 asks to identify the set of propaganda techniques used in a tweet, which is a multilabel classification problem, while Subtask~2 asks to detect the propaganda techniques used in a tweet together with the exact span(s) of text in which each propaganda technique appears. The task attracted 63 team registrations, and eventually 14 and 3 teams made submissions for subtask 1 and 2, respectively. Finally, 11 teams submitted system description papers. 
\end{abstract}

\section{Introduction}
\label{sec:introduction}

Social media platforms have become an important communication channel, where we can share and access information from a variety of sources. Unfortunately, the rise of this democratic information ecosystem was accompanied by and dangerously polluted with misinformation, disinformation, and malinformation in the form of propaganda, conspiracies, rumors, hoaxes, fake news, hyper-partisan content, falsehoods, hate speech, cyberbullying, etc. ~\cite{oshikawa-etal-2020-survey,alam2020fighting,pramanick-etal-2021-detecting,rosenthal-etal-2021-solid,alam-etal-2022-survey,FbMultiLingMisinfo:2022,guo-etal-2022-survey,hardalov-etal-2022-survey,10.1145/3517214,ijcai2022p781} 

Propaganda is conveyed through the use of diverse propaganda techniques~\cite{Miller}, which range from leveraging on the emotions of the audience (e.g., using loaded language, appealing to fear, etc.) to using logical fallacies such as \textit{straw men} (misrepresenting someone's opinion), \textit{whataboutism}, \textit{red herring} (presenting irrelevant data), etc. In the last decades, propaganda was widely used on social media to influence and/or mislead the audience, which became a major concern for different stakeholders, social media platforms, and policymakers. To address this problem, the research area of \emph{computational propaganda} has emerged, and here we are particularly interested in automatically identifying the use of propaganda techniques in text, images, and multimodal content. Prior work in this direction includes identifying propagandistic content in an article based on writing style and readability level~\cite{rashkin-EtAl:2017:EMNLP2017,BARRONCEDENO20191849}, at the sentence and the fragment levels from news articles with fine-grained techniques~\cite{EMNLP19DaSanMartino}, and in memes
\cite{dimitrov2021detecting}. 
These efforts focused on English, and there was no prior work on Arabic. Our shared task aims to bridge this gap by focusing on detecting propaganda in Arabic social media text, i.e., tweets.

\section{Related Work}
\label{sec:related_work}

In the current information ecosystem, propaganda has evolved to \textit{computational propaganda} \cite{woolley2018computational,da2020survey}, where information is distributed on social media platforms, which makes it possible for malicious users to reach well-targeted communities at high velocity. Thus, research on propaganda detection has focused on analyzing not only news articles but also social media content \cite{rashkin-EtAl:2017:EMNLP2017,BARRONCEDENO20191849,EMNLP19DaSanMartino,da2020survey,RANLP2021:COVID19:Bulgarian,RANLP2021:COVID19:Qatar,Propaganda:Coordination:WebSci:2022}. 

\citet{rashkin-EtAl:2017:EMNLP2017} focused on article-level propaganda analysis. They developed the \rashkincorpus~corpus, which used distant supervision for annotation with four classes: \emph{trusted}, \emph{satire}, \emph{hoax}, and \emph{propaganda}. The assumption of their distant supervision approach was that all articles from a given news source should share the same label. They collected their articles from the English Gigaword corpus and from seven other unreliable news sources, including two propagandistic ones. 
Later, \citet{BARRONCEDENO20191849} developed a new corpus, \proppycorpus, with two labels: propaganda vs. non-propaganda, and also experimented on \rashkincorpus~and \proppycorpus~corpora. For the \rashkincorpus~ corpus, they binarized the labels: propaganda \textit{vs.} any of the other three categories as non-propaganda. They investigated the writing style and the readability level of the target document, and trained models using logistic regression and SVMs. Their findings confirmed that using distant supervision, in conjunction with rich representations, might encourage the model to predict the source of the article, rather than to discriminate propaganda from non-propaganda. Similarly, \citet{Habernal.et.al.2017.EMNLP,Habernal2018b} developed a corpus with 1.3k arguments annotated with five fallacies, including \textit{ad hominem}, \textit{red herring}, and \textit{irrelevant authority}, which directly relate to propaganda techniques.


Recently, \citet{EMNLP19DaSanMartino}, curated a set of persuasive techniques, ranging from leveraging on the emotions of the audience such as using \textit{loaded language} and \textit{appeal to fear}, to logical fallacies such as \textit{straw man} (misrepresenting someone's opinion) and \textit{red herring} (presenting irrelevant data). They focused on textual content, i.e., newspaper articles. In particular, they developed a corpus of news articles annotated with eighteen propaganda techniques. The annotation was at the fragment level, and could be used for two tasks: (\emph{i})~binary classification ---given a sentence in an article, predict whether any of the 18 techniques has been used in it, and (\emph{ii})~multi-label classification and span detection task ---given a raw text, identify both the specific text fragments where a propaganda technique is used as well as the specific technique. They further proposed a multi-granular deep neural network that captures signals from the sentence-level task and helps to improve the fragment-level classifier. 
\citet{da2020semeval} also organized a shared task on Detection of Propaganda Techniques in News Articles. 

Subsequently, \citet{SemEval2021-6-Dimitrov} 
organized the SemEval-2021 task 6 on Detection of Propaganda Techniques in Memes. It had a multimodal setup, combining text and images, and asked participants to build systems to identify the propaganda techniques used in a given meme.
\citet{yu-etal-2021-interpretable} looked into interpretable propaganda detection.

Other related shared tasks include the FEVER task~\cite{thorne-etal-2018-fever} on fact extraction and verification, the Fake News Challenge~\cite{hanselowski-etal-2018-retrospective}, the FakeNews task at MediaEval~\cite{pogorelov2020fakenews}, as well as the NLP4IF tasks on propaganda detection \cite{da-san-martino-etal-2019-findings} and on fighting the COVID-19 infodemic in social media \cite{shaar2021findings}. Finally, we should mention the CheckThat! lab at CLEF~\cite{CheckThat:ECIR2019,clef-checkthat:2019,clef-checkthat-lncs:2020,clef-checkthat-en:2020,clef-checkthat-ar:2020,CheckThat:ECIR2021,clef-checkthat:2021:LNCS,clef-checkthat:2021:task1,clef-checkthat:2022:task1,ECIR:CLEF:2022,clef-checkthat:2022:LNCS,clef-checkthat:2022:task2}, which addresses many aspects of disinformation for different languages over the years such as fact-checking, verifiable factual claims, check-worthiness, attention-worthiness, and fake news detection.

The present shared task is inspired from prior work on propaganda detection. In particular, we adapted the annotation instructions and the propaganda techniques discussed in \cite{EMNLP19DaSanMartino,SemEval2021-6-Dimitrov}. 

\begin{figure*}[tbh!]
    \centering
    \caption{An example of tweet annotation with propaganda techniques \textit{loaded language} and \textit{name calling}.
    \label{fig:annotation_example_task_id_7065}
    }    
    \includegraphics[scale=0.35]{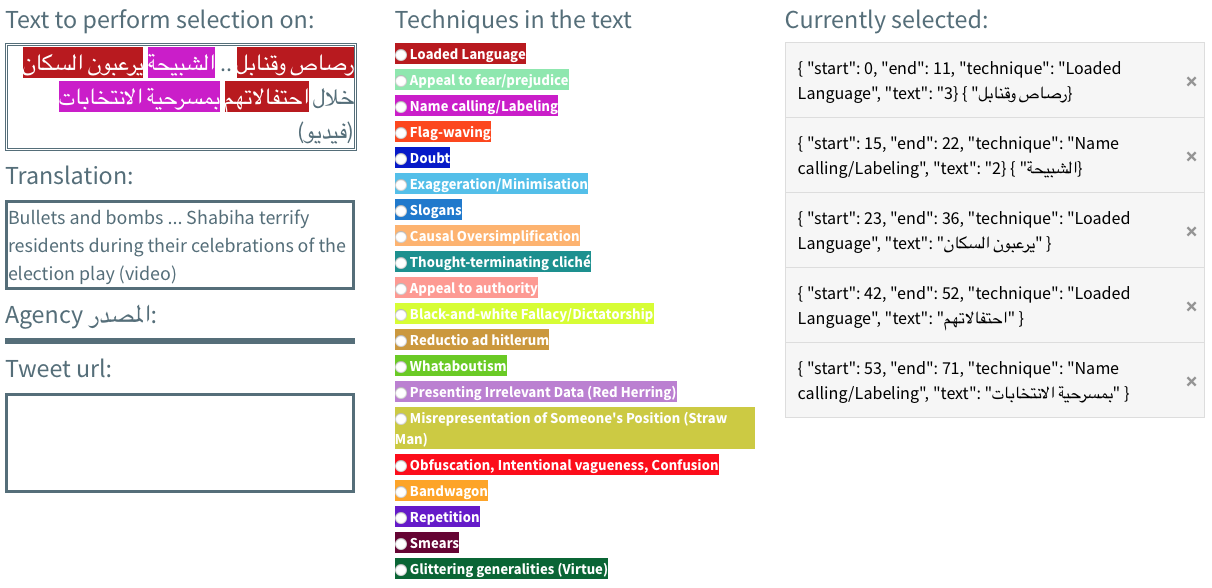}
\end{figure*}

\begin{figure*}[tbh!]
    \centering
    \caption{An example of tweet annotation with propaganda techniques \textit{loaded language} and \textit{slogan}.
    \label{fig:annotation_example_task_id_7922}
    }    
    \includegraphics[scale=0.35]{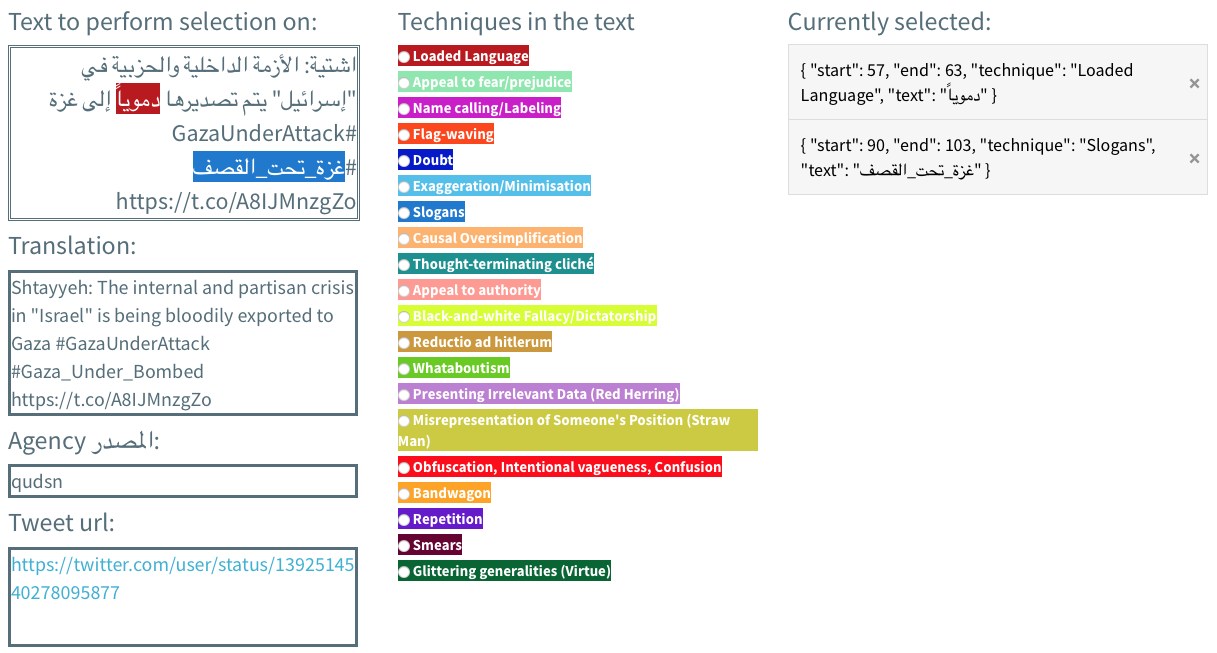}
\end{figure*}

\section{Tasks and Dataset}
\label{sec:task_dataset}

Below, we first formulate the two subtasks of our shared task, and then we discuss our datasets, including how we collected the data and what annotation guidelines we used.

\subsection{Tasks}
In the shared tasks, we offered the following two subtasks:

\begin{itemize}
    \item \textbf{Subtask 1:} Given the text of a tweet, identify the propaganda techniques used in it. 
    \item \textbf{Subtask 2:} Given the text of a tweet, identify the propaganda techniques used in it together with the span(s) of text in which each propaganda technique appears. 
\end{itemize}

Note that Subtask~1 is formulated as a multilabel classification problem, while Subtask~2 is a sequence labeling task. 

\subsection{Dataset}
We used Social Bakers\footnote{\url{https://www.socialbakers.com/}} to obtain the top-2 news sources from each Arab country, e.g.,~Al Arabiya and Sky News Arabia from UAE, Al Jazeera and Al Sharq from Qatar, etc. We further added five international sources that broadcast Arabic news: Al-Hurra News, BBC Arabic, CNN Arabic, France 24, and Russia Today. We then extracted from Twitter their latest 3,200 tweets. To have a balanced dataset that covers a wide range of topics, we chose 100 random tweets from each source, and then we sampled 930 tweets for annotation.

We target emotional appeals (e.g.,~loaded language, appeal to fear, flag waving, exaggeration, etc.) and logical fallacies (e.g.,~whataboutism, causal oversimplification, red herring, band wagon, etc.). We adopted the same techniques studied in \cite{EMNLP19DaSanMartino,SemEval2021-6-Dimitrov}. Below we briefly summarize them:
\begin{enumerate}
    \item \textbf{Appeal to authority:} Stating that a claim is true simply because a valid authority or expert on the issue said it was true. We also include here the special case where the reference is not an authority or an expert, which is referred to as \emph{Testimonial} in the literature.
    \item \textbf{Appeal to fear / prejudices:} Seeking to build support for an idea by instilling anxiety and/or panic in the population towards an alternative. In some cases, the support is built based on preconceived judgements.
    \item \textbf{Bandwagon} Attempting to persuade the target audience to join in and take the course of action because ``everyone else is taking the same action.''

    \item \textbf{Black-and-white fallacy or dictatorship:} Presenting two alternative options as the only possibilities, when in fact more possibilities exist. As an the extreme case, tell the audience exactly what actions to take, eliminating any other possible choices (ictatorship).
    \item \textbf{Causal oversimplification:} Assuming a single cause or reason when there are actually multiple causes for an issue. This includes transferring blame to one person or group of people without investigating the complexities of the issue.
    \item \textbf{Doubt:} Questioning the credibility of someone or something.       
    \item \textbf{Exaggeration / minimisation:} Either representing something in an excessive manner: making things larger, better, worse (e.g., \emph{the best of the best}, \emph{quality guaranteed}) or making something seem less important or smaller than it really is (e.g.,~saying that an insult was actually just a joke).
        
    \item \textbf{Flag-waving:} Playing on strong national feeling (or to any group, e.g.,~race, gender, political preference) to justify or to promote an action or an idea.
    
    \item \textbf{Glittering generalities (virtue)} These are words or symbols in the value system of the target audience that produce a positive image when attached to a person or issue. Peace, hope, happiness, security, wise leadership, freedom, ``The Truth'', etc. are virtue words. Virtue can be also expressed in images, where a person or an object is depicted positively.
    
    \item \textbf{Loaded language:} Using specific words and phrases with strong emotional implications (either positive or negative) to influence an audience.
    
    \item \textbf{Misrepresentation of someone's position (straw man):} Substituting an opponent's proposition with a similar one, which is then refuted in place of the original proposition.
    
    \item \textbf{Name calling or labeling:} Labeling the object of the propaganda campaign as something that the target audience fears, hates, finds undesirable or loves, praises.
    
    \item \textbf{Obfuscation, intentional vagueness, confusion:} Using words that are deliberately not clear, so that the audience may have their own interpretations. For example, when an unclear phrase with multiple possible meanings is used within an argument and, therefore, it does not support the conclusion.
    
    \item \textbf{Presenting irrelevant data (red herring):} Introducing irrelevant material to the issue being discussed, so that everyone's attention is diverted away from the points made.
    
    \item \textbf{Reductio ad hitlerum:} Persuading an audience to disapprove an action or an idea by suggesting that the idea is popular with groups hated in contempt by the target audience. It can refer to any person or concept with a negative connotation.
    
    \item \textbf{Repetition:} Repeating the same message over and over again, so that the audience will eventually accept it.
    
    \item \textbf{Slogans:} A brief and striking phrase that may include labeling and stereotyping. Slogans tend to act as emotional appeals.
    
    \item \textbf{Smears} A smear is an effort to damage or call into question someone's reputation, by propounding negative propaganda. It can be applied to individuals or groups.

    \item \textbf{Thought-terminating clich\'{e}:} Words or phrases that discourage critical thought and meaningful discussion about a given topic. They are typically short, generic sentences that offer seemingly simple answers to complex questions or that distract the attention away from other lines of thought.
    
    \item \textbf{Whataboutism:} A technique that attempts to discredit an opponent's position by charging them with hypocrisy without directly disproving their argument.

\end{enumerate}

\begin{table}[]
\centering
\caption{Statistics about the corpus. In parentheses, we show the number of tweets. \emph{Total} represents the number of techniques in each set.}
\label{tab:data_dist}
\scalebox{0.65}{
\setlength{\tabcolsep}{2.5pt}
\begin{tabular}{@{}lrrrr@{}}
\toprule
\multicolumn{1}{c}{\textbf{Prop Technique}} & \multicolumn{1}{c}{\begin{tabular}[c]{@{}c@{}}\textbf{Train}\\\textbf{(504)}\end{tabular}} & \multicolumn{1}{c}{\begin{tabular}[c]{@{}c@{}}\textbf{Dev}\\\textbf{(52)}\end{tabular}} & \multicolumn{1}{c}{\begin{tabular}[c]{@{}c@{}}\textbf{Dev-Test}\\\textbf{(51)}\end{tabular}} & \multicolumn{1}{c}{\begin{tabular}[c]{@{}c@{}}\textbf{Test}\\\textbf{(323)}\end{tabular}} \\ \midrule
 
Appeal to authority & 21 & 7 & 1 & 1 \\
Appeal to fear/prejudice & 48 & 7 & 4 & 25 \\
Black-and-white Fallacy/Dictatorship & 2 & 1 & 2 & 7 \\
Causal Oversimplification & 4 & 1 & 1 & 4 \\
Doubt & 29 & 1 & 2 & 19 \\
Exaggeration/Minimisation & 44 & 10 & 16 & 26 \\
Flag-waving & 5 & 2 & 2 & 9 \\
\begin{tabular}[c]{@{}l@{}}Glittering generalities \\
(Virtue)\end{tabular} & 25 & 7 & 2 & 1 \\
Loaded Language & 446 & 46 & 42 & 326 \\
Name calling/Labeling & 244 & 44 & 33 & 163 \\
\begin{tabular}[c]{@{}l@{}}Obfuscation, Intentional \\
vagueness, Confusion\end{tabular}& 9 & 3 & 1 & 6 \\
Presenting Irrelevant Data (Red Herring) &  1 & 0 & 0 & 0 \\
Repetition & 9 & 2 & 1 & 3 \\
Slogans & 44 & 1 & 1 & 6 \\
Smears & 85 & 12 & 15 & 50 \\
Thought-terminating cliché & 6 & 1 & 1 & 0 \\
Whataboutism & 3 & 1 & 1 & 0 \\ \cmidrule{2-5}
\textbf{Total} & \textbf{1025} & \textbf{146} & \textbf{125} & \textbf{646} \\
\bottomrule
\end{tabular}
}

\end{table}

The annotation is done in different stages: (\emph{i})~three annotators independently annotate the same tweet, and (\emph{ii})~they meet together with one consolidator to discuss each instance and to come up with gold annotations. Since the annotations are at the fragment level, it might happen that an annotation is spotted by only one annotator. The two phases ensure that each annotation is eventually discussed by all annotators. In order to train the annotators, we provide clear annotation instructions with examples and ask them to annotate a sample of tweets. Then, we revise their annotations and provide feedback. Figures \ref{fig:annotation_example_task_id_7065} and \ref{fig:annotation_example_task_id_7922} show example tweets with annotated propaganda techniques.  

Table \ref{tab:data_dist} shows the distribution of the propaganda techniques in our dataset for different data splits. Our annotation guidelines inclide twenty techniques, but in the annotated dataset, there were no instances of \emph{bandwagon}, \emph{straw man}, and \emph{reductio ad hitlerum}. Overall, the distribution of the propaganda techniques in our dataset is very skewed, which made the task challenging.

\section{Evaluation Framework}
\label{sec:evaluation}

\subsection{Evaluation Measures}
\label{ssec:evaluation_measures}
To measure the performance of the systems, for both subtasks, we use micro-F1 and macro-F1, as these are multi-class multi-label problems, where the labels are imbalanced. The official evaluation measure for subtask 1 is micro-F1, but the scorer also reports macro-F1. 

Subtask 2 is a multi-label sequence tagging problem. We modified the standard micro-averaged F1 to account for partial matching between the spans. More details about the modified macro-averaged F1 can be found in \cite{EMNLP19DaSanMartino,SemEval2021-6-Dimitrov}.

\subsection{Task Organization}
\label{task_organization}

We ran the shared task in two phases:

\paragraph{Development Phase} In the first phase, we provided the participants three subsets of the dataset: train, dev, and dev\_test. The purpose of the dev set was to fine-tune the trained model, and the dev\_test set was to evaluate the model performance on unseen dev\_test set. 

\paragraph{Test Phase} In the second phase, we released the actual test set and the participants were given just a few days to submit their final predictions via the submission system on Codalab.\footnote{\url{https://codalab.lisn.upsaclay.fr/competitions/7274}} 
In this phase, the participants could again submit multiple runs, but they would not get any feedback on their performance. Only the latest submission of each team was considered as official and was used for the final team ranking. The final leaderboard on the test set was made publicly available after the system submission deadline. 

\section{Participants and Results}
\label{sec:results}
In this section, we provide a general description of the systems that participated in each subtask and their results. Table~\ref{tab:results_subtasks} shows the results for all teams for both subtasks, as well as a random baseline. We can see that subtask 1 was more popular, attracting submissions by 14 teams, while there were only three submissions for subtask 2.

\begin{table}[]
\centering
\caption{Results for subtask 1 on multilabel propaganda detection and subtask 2 on identifying propaganda techniques and their 
span(s) in the text. The results are ordered by the official score: Micro-F1. $^*$Indicated that no system description paper was submitted.}
\label{tab:results_subtasks}
\scalebox{0.62}{
\setlength{\tabcolsep}{2.5pt}
\begin{tabular}{@{}lrr@{}}
\toprule
\multicolumn{1}{c}{\textbf{Rank/Team}} & \multicolumn{1}{c}{\textbf{Macro F1}} & \multicolumn{1}{c}{\textbf{Micro F1}} \\ \midrule
\multicolumn{3}{c}{\textbf{Subtask 1}} \\ \midrule
1. NGU\_CNLP~\cite{wanlp:2022:task1_2:Samir} & 0.185 & 0.649 \\
2. IITD~\cite{wanlp:2022:task1_2:iitd} & 0.183 & 0.609 \\
3. CNLP-NITS-PP~\cite{wanlp:2022:task1:CNLP-NITS-PP} & 0.068 & 0.602 \\
3. AraBEM~\cite{wanlp:2022:task1:AraBEM} & 0.068 & 0.602 \\
3. Pythoneers~\cite{wanlp:2022:task1_2:attieh} & 0.177 & 0.602 \\
4. AraProp~\cite{wanlp:2022:task1:GSingh} & 0.105 & 0.600 \\
5. iCompass~\cite{wanlp:2022:task1:Taboubi} & 0.191 & 0.597 \\
6. SI2m \& AIOX Labs~\cite{wanlp:2022:task1:Gaanoun} & 0.137 & 0.585 \\
7. mostafa-samir$^*$ & 0.186 & 0.580 \\
8. Team SIREN AI~\cite{wanlp:2022:task1:SIREN_AI} & 0.153 & 0.578 \\
9. ChavanKane~\cite{wanlp:2022:task1:Chavan} & 0.111 & 0.565 \\
10. mhmud.fwzi$^*$ & 0.087 & 0.552 \\
11. TUB~\cite{wanlp:2022:task1:mohtaj} & 0.076 & 0.494 \\
12. tesla$^*$ & 0.120 & 0.355 \\
13.  \textit{Baseline (Random)} & 0.043 & 0.079 \\ \midrule
\multicolumn{3}{c}{\textbf{Subtask 2}} \\ \midrule
1. Pythoneers~\cite{wanlp:2022:task1_2:attieh} & & 0.396 \\
2. IITD~\cite{wanlp:2022:task1_2:iitd} & & 0.355 \\
3. NGU\_CNLP~\cite{wanlp:2022:task1_2:Samir} & & 0.232 \\
4. \textit{Baseline (Random)}& & 0.013 \\
\bottomrule
\end{tabular}
}

\end{table}


\subsection{Subtask 1}

Table \ref{tab:overview_subtasks} gives an overview of the systems that took part in subtask 1. We can see that transformers were quite popular, most notably AraBERT, followed by BERT, and MARBERT. Some participants also 
used ensembles methods, data augmentation, and standard preprocessing. 

The best-performing team NGU\_CNLP \cite{wanlp:2022:task1_2:Samir} first explored various baselines models such as bag of words with SVM, Na\"{i}ve Bayes, Stochastic Gradient Descent, Logistic Regression, Random Forests and K-nearest Neighbor. Eventually, for their final submission, they used AraBERT with stacking-based ensemble (5-fold split). They further explored translation-based data augmentation using the English PTC corpus \cite{EMNLP19DaSanMartino}.

The second best system was IITD~\cite{wanlp:2022:task1_2:iitd}, and they used XLM-R and fine-tuned the model. They also explored data augmentation by translating ad adding the PTC corpus as training, but in their experiments this did not help improve the performance. 

The third system was CNLP-NITS-PP~\cite{wanlp:2022:task1:CNLP-NITS-PP}, and they used the AraBERT Twitter-base model along with data augmentation. Note that all systems outperformed the random baseline. 

\subsection{Subtask 2}
In Table \ref{tab:overview_subtasks}, we also present an overview of the systems that took part in Subtask 2. Once again, this subtask was dominated by transformer models. We can see in the table that transformers were quite popular, and among them, the most commonly used one was AraBERT, followed by BERT and MARBERT. The participants in this task also used data augmentation and standard pre-processing. 

Table~\ref{tab:results_subtasks} shows the evaluation results: we report our random baseline, which is based on the random selection of spans with random lengths and a random assignment of labels. 

The best system for this subtask was Pythoneers~\cite{wanlp:2022:task1_2:attieh}. They used AraBERT with a Conditional Random Field (CRF) layer, which was trained on encoded data using the BIO schema. 

The second-best system was IITD~\cite{wanlp:2022:task1_2:iitd}, which used a Multi-Granularity Network \cite{EMNLP19DaSanMartino} with the mBERT encoder. 

The third system was NGU\_CNLP~\cite{wanlp:2022:task1_2:Samir}. They converted the data to BIO format and fine-tuned a token classifier based on Marefa-NER\footnote{\url{https://huggingface.co/marefa-nlp/marefa-ner}} (pretrained using XLM-RoBERTa).

\begin{table}[]
\centering
\caption{Overview of the approaches used for subtasks 1 and 2, for the teams that submitted a description paper. The systems are ordered by the official score: F1-micro.}
\label{tab:overview_subtasks}
\scalebox{0.58}{
\setlength{\tabcolsep}{2.5pt}
\begin{tabular}{@{}l|ccccc|ccc@{}}
\toprule
\multicolumn{1}{c|}{\textbf{Rank/Team}} & \multicolumn{5}{c|}{\textbf{Models}} & \multicolumn{3}{c}{\textbf{Other}} \\ \midrule
\multicolumn{1}{c|}{\textbf{}} & 
\rotatebox{90}{\textbf{BERT}} 
& \rotatebox{90}{\textbf{XML-R}} & \rotatebox{90}{\textbf{AraBERT}} & \rotatebox{90}{\textbf{ARBERT}} & \rotatebox{90}{\textbf{MARBERT}} & 
\rotatebox{90}{\textbf{Data augmentation}} & \rotatebox{90}{\textbf{Preprocessing}} & \rotatebox{90}{\textbf{NER}} \\ \midrule
\multicolumn{9}{c}{\textbf{Subtask 1}} \\  \midrule
1. NGU\_CNLP~\cite{wanlp:2022:task1_2:Samir} &  &  & \sq &  &  & \sq & \sq &  \\
2. IITD~\cite{wanlp:2022:task1_2:iitd} &  & \sq &  &  &  &  &  &  \\
3. CNLP-NITS-PP~\cite{wanlp:2022:task1:CNLP-NITS-PP} &  &  & \sq &  &  & \sq & \sq &  \\
3. AraBEM~\cite{wanlp:2022:task1:AraBEM} & \sq &  &  &  &  &  &  &  \\
3. Pythoneers~\cite{wanlp:2022:task1_2:attieh} &  &  & \sq &  &  &  & \sq &  \\
4. AraProp~\cite{wanlp:2022:task1:GSingh} &  &  &  &  & \sq &  & \sq &  \\
5. iCompass~\cite{wanlp:2022:task1:Taboubi} &  &  &  &  & \sq &  & \sq &  \\
6. SI2m \& AIOX Labs~\cite{wanlp:2022:task1:Gaanoun} &  &  &  & \sq &  & \sq &  & \sq \\
8. Team SIREN AI~\cite{wanlp:2022:task1:SIREN_AI} &  &  & \sq &  &  &  & \sq &  \\
9. ChavanKane~\cite{wanlp:2022:task1:Chavan} &  &  & \sq &  & \sq & \sq & \sq &  \\
11. TUB~\cite{wanlp:2022:task1:mohtaj} & \sq &  &  &  &  &  & \sq &  \\
\midrule
\multicolumn{9}{c}{\textbf{Subtask 2}} \\ \midrule
1. Pythoneers~\cite{wanlp:2022:task1_2:attieh} &  &  & \sq &  &  &  &  &  \\
2. IITD~\cite{wanlp:2022:task1_2:iitd} & \sq &  &  &  &  &  &  &  \\
3. NGU\_CNLP~\cite{wanlp:2022:task1_2:Samir} &  &  &  &  &  & \sq &  &  \\
\bottomrule
\end{tabular}
}

\end{table}

\subsection{Participants' Systems}

\paragraph{NGU\_CNLP \cite{wanlp:2022:task1_2:Samir}[\rank{subtask 1}{1}, \rank{subtask 2}{3}]} team participated in both subtasks. For subtask 1, they used a combination of a data augmentation strategy with a transformer-based model. This model ranked first among the 14 systems that participated in this subtask. Their preliminary experiments for subtask 1 consist of using a bag-of-words model with different classical algorithms such as Support Vector Machines, Naïve Bayes, Stochastic Gradient Descent, Logistic regression, Random Forests, and simple K-nearest Neighbor. For subtask 2, they fine-tuned the Marefa-NER model, which is based on XLM-RoBERTa. The system ranked third among the three systems that participated in this subtask.

\paragraph{Pythoneers \cite{wanlp:2022:task1_2:attieh}[\rank{subtask 1}{3}, \rank{subtask 2}{1}]} also participated in both subtasks. For subtask 1, they trained a multi-task learning model that performs binary classification per propaganda technique. For subtask 2, they first converted the data into BIO format and then fine-tuned an AraBERT model with a Conditional Random Field (CRF) layer. Their subtask 1 system ranked third with a micro-averaged F1-Score of 0.602, and their subtask 2 system ranked first with a micro-averaged F1-Score of 0.396.

\paragraph{IITD \cite{wanlp:2022:task1_2:iitd}[\rank{subtask 1}{2}, \rank{subtask 2}{2}]}. This team also participated in both subtasks. They used multilingual pretrained language models for both subtask s. For subtask 1, they used a pretrained XLM-R to estimate a Multinoulli distribution after projecting the CLS embedding to a 20-dimensional embedding (one per propaganda technique). 
For subtask 2, they used a multi-granularity network \cite{EMNLP19DaSanMartino} with mBERT encoder. Even though both systems were trained on only the dataset released in this shared task, they also discussed several methods (zero-shot transfer, continued training, and translation of PTC \cite{EMNLP19DaSanMartino} to {Arabic}) to study cross-lingual propaganda detection. This suggested interesting research challenges for future exploration, such as how to effectively use data from different domains and how to learn language-agnostic embeddings in propaganda detection systems.

\paragraph{CNLP-NITS-PP \cite{wanlp:2022:task1:CNLP-NITS-PP}[\rank{subtask 1}{3}].} This team participated in subtask 1 and they used AraBERT Twitter-base model for multilabel propaganda classification. They further used data augmentation; in particular, they generated synthetic training data using root and stem substitution from the original train samples and prepared additional synthetic examples. They changed the input labels to the model to be one-hot encoded to indicate multiple labels and modified the macro-F1 scorer to give a score for multiple labels. To make predictions with the model, they used a sentiment analysis pipeline from HuggingFace Transformers and selected all the labels that yielded a score greater than or equal to 0.32. They observed the scores for the predictions on the validation test set and found that most correct labels had a score greater than 0.30. They also found that there was a large gap in the score for the label when the score was below 0.30.

\paragraph{AraBEM \cite{wanlp:2022:task1:AraBEM}[\rank{subtask 1}{3}].} This team participated in subtask 1 and they fine-tuned BERT to perform multi-class binary classification. They used standard pre-processing including normalization (mapping letters with various forms, i.e., alef, hamza, and yaa to their representative characters), and removing special characters, diacritics, and repeated characters. 

\paragraph{AraProp \cite{wanlp:2022:task1:GSingh}[\rank{subtask 1}{4}].} This team participated in subtask 1. First, they tokenized the input and produced contextualized word embeddings for all input tokens. To get a fixed-size output representation, they simply averaged all contextualized word embeddings by taking attention mask into account for correct averaging. Then, they added a dropout layer with a dropout rate of 0.3, followed by a linear layer with a sigmoid activation function for the output. 
They experimented with multiple transformer-based language models: two multilingual models and six monolingual (Arabic) models. 
Their findings suggest that the {MARBERTv2}-based fine-tuned model outperforms other models in terms of F1-micro score.

\paragraph{iCompass \cite{wanlp:2022:task1:Taboubi}[\rank{subtask 1}{5}] } team participated in subtask 1. Their system used standard pre-processing such as normalization and removing stopwords, emojis, special characters, and links. Then, they used pre-trained language models such as MARBERT and ARBERT. They further added global average and max pooling layers on top of the models. Finally, they used cross-validation to improve the model performance.

\paragraph{SI2M \& AIOX Labs \cite{wanlp:2022:task1:Gaanoun}[\rank{subtask 1}{6}]} team participated in subtask 1. They used data augmentation, named entity recognition (NER), and manual rules. For data augmentation, they combined the training and the dev sets, and randomly mixed the sequences to create new synthetic sequences, which they concatenated with the train and the dev sets. Their final system uses a mixed dataset of 2,000 examples.
Next, they fine-tuned ARBERT on the augmented dataset, and they made predictions based on a defined threshold of the classifier's confidence.  
If no technique got a prediction probability greater than the threshold, the token was assigned the label \emph{No technique}. Moreover, to detect the \emph{Name Calling/Labelling} technique, they used a NER model based on AraBERT. 
Finally, to detect \emph{Repetition}, they used manual rules, after removing the stopwords.

\paragraph{Team SIREN AI \cite{wanlp:2022:task1:SIREN_AI}[\rank{subtask 1}{8}]} participated in subtask 1 and used AraBERT for fine-tuning. Like other teams, they used standard pre-processing, e.g.,~removing HTML markup, diacritics, non-digit repetitions, etc.

\paragraph{ChavanKane \cite{wanlp:2022:task1:Chavan}[\rank{subtask 1}{9}]} team participated in subtask 1 and experimented with AraBERT v1, v02 and v2, MARBERT, ARBERT, XLMRoBERTa, and AraELECTRA. They used a specific variant of DeHateBERT, which is initialized from multilingual BERT and fine-tuned only on Arabic datasets. They also tried creating an ensemble of all models, which consists of five models such as DeHateBERT, AraBERTv2, AraBERTv02, AraBERTv01, and MARBERT.
For the final prediction from the ensembles, they used hard voting.

\paragraph{TUB \cite{wanlp:2022:task1:mohtaj}[\rank{subtask 1}{11}].} This team participated in subtask 1 and used a semantic similarly detection approach based on conceptual word embedding. They converted all sentences in the train, dev, and test sets into vectors using the BERT model. For each sentence in the test set, they detected the five most similar instances from the train and the dev sets, with a cosine similarity above 0.4. Then, they assigned the three most frequent labels among the five instances as the label of the target sentence.

\section{Conclusion and Future Work}
\label{sec:conclusion}

We presented the WANLP'2022 shared task on \emph{Propaganda Detection in Arabic}, as part of which we developed the first dataset for Arabic propaganda detection with focus on social media content. This was a successful task: a total of 63 teams registered to participate, and 14 and 3 teams eventually made an official submission on the test set for subtasks 1 and 2, respectively. Finally, 11 teams submitted a task description paper. Subtask~1 asked to identify the propaganda techniques used in a tweet, and subtask~2 further asked to identify the the span(s) of text in which each propaganda technique appears. For both subtasks, the majority of the systems fine-tuned pre-trained Arabic language models, and used standard pre-processing. Some systems used data augmentation and ensemble methods.

In future work, we plan to increase the data size and to add hierarchically structured propaganda techniques.

\section{Acknowledgments}
This publication was made possible by NPRP grant
13S-0206-200281 \emph{Resources and Applications for Detecting and Classifying Polarized and Hate Speech in Arabic Social Media} from the Qatar National Research Fund. 

Part of this work was also funded by Qatar Foundation's IDKT Fund TDF 03-1209-210013: \emph{Tanbih: Get to Know What You Are Reading}.

This research is also carried out as part of the Tanbih mega-project,\footnote{\url{http://tanbih.qcri.org/}} developed at the Qatar Computing Research Institute, HBKU, which aims to limit the impact of ``fake news'', propaganda, and media bias, thus promoting digital literacy and critical thinking.

The findings herein are solely the responsibility of the authors.

\bibliography{bib/propaganda,bib/bibliography}
\bibliographystyle{acl_natbib}

\end{document}